\documentclass{article}

\usepackage{arxiv}
 
\usepackage[utf8]{inputenc} 
\usepackage[T1]{fontenc}    
\usepackage{hyperref}       
\usepackage{url}            
\usepackage{booktabs}       
\usepackage{amsfonts}       
\usepackage{nicefrac}       
\usepackage{microtype}      
\usepackage{lipsum}		
\usepackage{graphicx}
\usepackage{subcaption}
\usepackage{natbib}
\usepackage{doi}
\usepackage{float}
\title{SANA‑I2I: A Text‑Free Flow‑Matching Framework for Paired Image‑to‑Image Translation with a Case Study in Fetal MRI Artifact Reduction}

\author{ Santos I. M . F \\
	National Laboratory for Scientific Computing - LNCC \\
	\texttt{} \\
	\And 
	Werner H. \\
	Biodesign Laboratory Dasa \\
	\texttt{} \\
    \And 
    Giraldi A. G.\\
	National Laboratory for Scientific Computing - LNCC \\
	\texttt{} \\
}

\renewcommand{\shorttitle}{Sana-I2I}

\hypersetup{
pdftitle={A template for the arxiv style},
pdfsubject={q-bio.NC, q-bio.QM},
pdfauthor={David S.~Hippocampus, Elias D.~Striatum},
pdfkeywords={Image-to-image, flow matching, SANA, paired translation, high-resolution generation, Fetal MRI},
}

\begin{document}
\maketitle

\begin{abstract}
We propose \textbf{SANA-I2I}, a text-free, high-resolution image-to-image generation framework that extends the SANA family by removing textual conditioning entirely. In contrast to SanaControlNet, which combines text and image-based control, SANA-I2I relies exclusively on paired source--target images to learn a conditional flow-matching model in latent space. The model learns a conditional velocity field that maps a target image distribution to another one, enabling supervised image translation without reliance on language prompts.
We evaluate the proposed approach on the challenging task of fetal MRI motion artifact reduction. To enable paired training in this application, where real paired data are difficult to acquire, we adopt a synthetic data generation strategy based on the method proposed by Duffy et al., which simulates realistic motion artifacts in fetal magnetic resonance imaging (MRI).  Experimental results demonstrate that SANA-I2I effectively suppresses motion artifacts while preserving anatomical structure, achieving competitive performance few inference steps. These results highlight the efficiency and suitability of our proposed flow-based, text-free generative models for supervised image-to-image tasks in medical imaging.
\end{abstract}

\section{Introduction}
High-resolution image generation has advanced rapidly with the development of diffusion and flow-based generative models, leading to substantial improvements in visual fidelity, scalability, and controllability \cite{ho2020ddpm, song2020score, dhariwal2021diffusion, lipman2023flow}. Among recent approaches, SANA \cite{sana2024} introduces an efficiency-oriented framework for text-to-image generation based on Flow Matching, achieving competitive high-resolution image synthesis with significantly reduced computational cost compared to conventional diffusion models. By jointly optimizing the training objective, network architecture, and inference procedure, SANA demonstrates that flow-based generative modeling is a viable and efficient alternative for large-scale image synthesis.

Beyond text-to-image generation, SANA was extended to SanaControlNet \cite{sana2024}, which enables image-conditioned generation by combining a textual prompt with a structural image input. In this formulation, the conditioning image (e.g., edges, sketches, or layouts) provides spatial guidance that is injected into the generative backbone, while text embeddings supply semantic constraints. This design follows principles introduced in ControlNet \cite{zhang2023controlnet}, allowing the model to preserve spatial structure derived from the image while remaining aligned with textual semantics. However, the reliance on text conditioning introduces practical and conceptual limitations in scenarios where text annotations are unavailable, ambiguous, or unnecessary, as is often the case in image-to-image translation problems.

In many real-world image transformation tasks, such as denoising, artifact removal, or modality translation, the desired output is fully specified by an input image \cite{zhang2017dncnn,saharia2022palette, chartsias2017multimodal}. In these settings, paired datasets consisting of aligned source and target images are either naturally available or can be reliably constructed, for example through controlled acquisition protocols or an artificial corruption processes. For such problems, textual conditioning offers little additional information and may introduce unwanted stochasticity or inconsistencies that conflict with the nature of image-to-image translation tasks. Despite this, most high-fidelity diffusion or flow-based image translation methods remain tightly coupled to text prompts, often adapting text-to-image backbones rather than explicitly modeling paired image transformations.

To address this gap, we introduce a text-free image-to-image formulation within the SANA's family framework, explicitly designed for supervised paired image translation. Our approach removes textual conditioning entirely and instead learns a conditional generative mapping driven solely by an input image, using Flow Matching \cite{lipman2023flow} methodology in latent space. Given paired source–target images, the model is trained to learn a conditional velocity field that transforms noise into a target image or latent distribution while being guided exclusively by the source image. This formulation preserves the efficiency and stability of SANA’s flow-based training while aligning the model more naturally with image-to-image tasks.

Conceptually, our method can be viewed as a specialization of SanaControlNet \cite{sana2024} in which the input image serves as the only conditioning signal, responsible for both spatial structure and content constraints. Rather than complementing text with structural guidance, we directly condition the generative process on image-derived features at multiple resolutions. Importantly, we retain the classifier-free guidance (CFG) mechanism, originally introduced for diffusion models \cite{ho2022classifierfree}, which allows explicit control over the strength of the image transformation at inference time without retraining. 

To enable supervised training under this formulation, we adopt synthetic data generation strategies to produce aligned source–target pairs. In the context of fetal MRI artifact removal, such synthetic generation provides a practical way to balance artifact suppression against the preservation of fine anatomical details, allowing clinicians to adjust the aggressiveness of the correction according to diagnostic needs. Specifically, we employ a strategy proposed by Duffy et al. \cite{Duffy2021RetrospectiveMA}, in which realistic motion artifacts are artificially introduced into high‑quality images to create paired training data. Clean images serve as ground truth, while their degraded counterparts are generated through controlled corruption processes designed to mimic real acquisition artifacts. This approach enables the construction of large‑scale paired datasets with precise pixel‑level alignment, without requiring repeated acquisitions or manual annotation. Although training relies on these synthetic degradations, the model is evaluated on both synthetic and real fetal MRI data to assess its robustness and generalization to real‑world imaging distributions.

The experimental results show that the text-free SANA image-to-image model produces high-quality reconstructions that preserve anatomical structure while effectively reducing motion artifacts. Moreover, the proposed formulation achieves strong performance with even with few steps, highlighting the computational efficiency of flow-based image-to-image generation.

In summary, this work introduces the following key contributions:
\begin{itemize}
    \item We propose a text-free image-to-image extension of SanaControlNet tailored for supervised paired image translation.
    \item We show that removing textual conditioning better aligns the model with image transformation tasks and substantially reduces the number of inference steps required in the considered application.
    \item We demonstrate that the proposed approach preserves the efficiency and scalability advantages of SANA's family for high-resolution generation.
    \item We validate the method on the challenging task of fetal MRI motion artifact reduction, highlighting its potential for other applications in medical imaging.
\end{itemize}

The following text is organized as follows: we begin by reviewing the related works in Section \ref{sec:related-works}, which contextualize the proposed approach within existing image-to-image methodologies and flow-matching frameworks. Section \ref{sec:dados} then presents the data used in this study, including both the synthetic paired dataset and the real fetal MRI acquisitions. The proposed methodology is introduced in Section \ref{sec:methodology}, detailing the architectural design, conditioning mechanisms, and training strategy. Section \ref{sec:training-configs} describes the training configurations, the parameters chosen and ablation settings. The experimental results are reported and analyzed in Section \ref{sec:results}, including comparisons with previous methods and quantitative, qualitative, and distribution-based evaluations on both synthetic and real data. Finally, Section \ref{sec:final-remarks} provides the final remarks, summarizing the main findings and highlighting the broader applicability and potential extensions of the proposed adaptation.

\section{Related Works} \label{sec:related-works}

\subsection{Diffusion and Flow-Based Generative Models}
Diffusion models have emerged as a dominant paradigm for high-fidelity image synthesis \cite{ho2020ddpm, song2020score, dhariwal2021diffusion}, enabling state-of-the-art results across text-to-image and image-to-image generation tasks. More recently, Flow Matching (FM) \cite{lipman2023flow} has been introduced as an alternative continuous-time formulation that directly estimates the velocity field between data and noise distributions, offering improved training stability and reduced sampling complexity compared to classical diffusion processes. Building upon FM, SANA \cite{sana2024} demonstrates that flow-based models can scale to high resolutions while maintaining computational efficiency through a combination of architectural optimizations, latent-space compression, and tailored training objectives.

\subsection{Image-Conditioned Generation and Control Mechanisms}
Image-conditioned generative modeling has been extensively studied in the context of controllable synthesis. ControlNet \cite{zhang2023controlnet} popularized the injection of spatial conditions, such as edges, segmentation masks, or sketches, into diffusion backbones, enabling enhanced structural fidelity. SanaControlNet \cite{sana2024} extended this paradigm to the SANA framework by integrating a ControlNet-like branch with textual conditioning. Although highly effective for structured conditional generation, these approaches inherently rely on text prompts, which may be unnecessary or detrimental in tasks where conditioning images fully specify the output domain.

In contrast, a long line of paired image-to-image (I2I) translation work, such as Pix2Pix \cite{isola2017pix2pix}, supervised denoising frameworks \cite{zhang2017dncnn}, and modality-translation networks, demonstrates that many real-world transformations can be accurately learned from paired images alone, without semantic textual guidance. Recent diffusion-based I2I approaches, such as DiffI2I \cite{xia2023diffi2i}, the Brownian Bridge Diffusion Model (BBDM) \cite{li2023bbdm}, and the Diffusion Model Translator (DMT) \cite{xia2024dmt}, introduce efficient translation mechanisms and improved domain-transfer strategies, further advancing paired and conditional formulations. Our formulation aligns with these supervised I2I trends by removing text entirely and leveraging only image-derived latent representations.

A recent contribution particularly relevant to text-free, paired image-to-image translation is the Latent Bridge Matching (LBM) framework \cite{chadebec2025lbm}. LBM introduces a latent-space bridge-matching formulation that enables fast image-to-image translation, achieving state-of-the-art performance across tasks such as object removal, normal and depth estimation, and relighting, often with only a single inference step. Notably, the adapted SANA-I2I model also produces high-quality results using a minimal number of sampling steps.

\subsection{Motion Artifact Simulation and Fetal MRI Correction}
Motion artifacts remain a major challenge in fetal MRI due to unpredictable fetal and maternal motion. To address the lack of paired clean/corrupted datasets, several works have introduced retrospective artifact simulation techniques. In particular, Duffy et al. \cite{Duffy2021RetrospectiveMA} proposed a realistic k-space motion simulation method that has been adopted for MR motion-correction tasks \cite{RetroMRMotionCorrection}. Deep-learning-based fetal MRI reconstruction methods typically rely on unpaired adversarial learning \cite{santos2024fetal,Lim2023MotionAC} or self-supervised techniques, but these approaches struggle to maintain anatomical fidelity because some harmful artifacts are not removed in the image or insufficient results are provided for high level of artifacts.

Our work builds upon this foundation by adopting the Duffy simulator to generate aligned paired data, enabling supervised training of a flow-based generative model for artifact correction. Unlike SanaControlNet, our adaptation eliminates textual conditioning entirely, allowing the model to focus on structural consistency and domain-specific transformations relevant to medical imaging.

\section{Methodology}
\label{sec:methodology}

We propose a text-free image-to-image (I2I) extension of SANA, formulated as a conditional flow-matching model operating in latent space. The proposed method is derived from SanaControlNet \cite{sana2024}, but removes textual conditioning entirely while preserving both the control architecture and the classifier-free guidance (CFG) mechanism originally introduced in \cite{ho2022classifierfree}.  The central idea is to reinterpret the conditioning variable $y$, originally representing text embeddings in SANA, as an image-based or latent signal, enabling image-conditioned generation without semantic text input.

We introduce two variants of the proposed framework:
\begin{itemize}
    \item \textbf{Primary approach}: only the conditioning variable $y$ is randomly dropped for CFG, while the control pathway remains active.
    \item \textbf{Bis (exploratory) approach}: both $y$ and the control signal are jointly dropped, enabling fully unconditional generation.
\end{itemize}
The Bis variant is included to highlight the flexibility of the proposed formulation: while the primary approach is better suited for paired I2I tasks, where strong structural fidelity and accurate artifact removal are essential, the bis approach additionally provides the ability to generate images directly from the target domain without any conditioning input. This makes the second variant particularly valuable when both conditional and unconditional generation capabilities are desired within a single unified model. Such hybrid conditional/unconditional systems have been studied in prior generative modeling frameworks \cite{ho2022classifierfree}. Nevertheless, unless otherwise stated, this paper focuses on the primary approach, which empirically yields superior performance for fetal artifact removal.

Following SANA, we adopt the Deep Compression Autoencoder (DC-AE) \cite{chen2025dcae}, which was introduced as a high-compression latent autoencoder designed to accelerate high‑resolution diffusion and flow-based models by encoding images into a compact latent representation and decode them back to image space. DC-AE enables aggressive spatial compression while maintaining high reconstruction fidelity, substantially reducing the number of tokens processed by the generative model at high resolutions. This design choice aligns with SANA’s emphasis on efficient high-resolution generation. In this work, we use the same DC-AE model provided in the original SANA implementation.

\subsection{From SanaControlNet to Text-Free SANA-I2I}

SanaControlNet models the conditional process using the tuple $(x, t, y, control)$, where $x$ denotes Gaussian noise (or an intermediate latent state), $t$ is the continuous time variable, $y$ represents a text embedding, and $control$ is an image or latent-based signal injected through a ControlNet branch to provide spatial guidance \cite{sana2024}. This formulation closely follows the principles of ControlNet \cite{zhang2023controlnet}, where spatial conditioning complements semantic conditioning provided by text. In the original SANA formulation, textual information in $y$ is encoded using a dedicated CaptionEmbed module, while both $x$ and the control signal are processed through PatchEmbed networks.

In the proposed text-free SANA-I2I formulation, we remove textual conditioning entirely and redefine the model inputs as $(x, t, y)$, where $y$ now denotes an image or a latent representation used as the conditioning signal. For architectural compatibility, we set $y \equiv control$, meaning that the same image-based latent simultaneously fulfills the roles of conditioning and spatial guidance.

Unlike the original SANA, where the text conditioning is processed by a CaptionEmbed, in our formulation, both $y$ and the control signal are embedded using the same PatchEmbed network employed for $x$. This design choice allows all modalities noise, conditioning, and control to reside in the same latent space and share a unified embedding mechanism, while preserving the original ControlNet structure and feature injection points. In contrast to SanaControlNet, which already applies PatchEmbed to the control signal but retains a separate embedding pathway for $y$, our implementation consolidates these embeddings into a single shared PatchEmbed network.

\begin{figure}[H]
    \centering
    \includegraphics[width=0.8\linewidth]{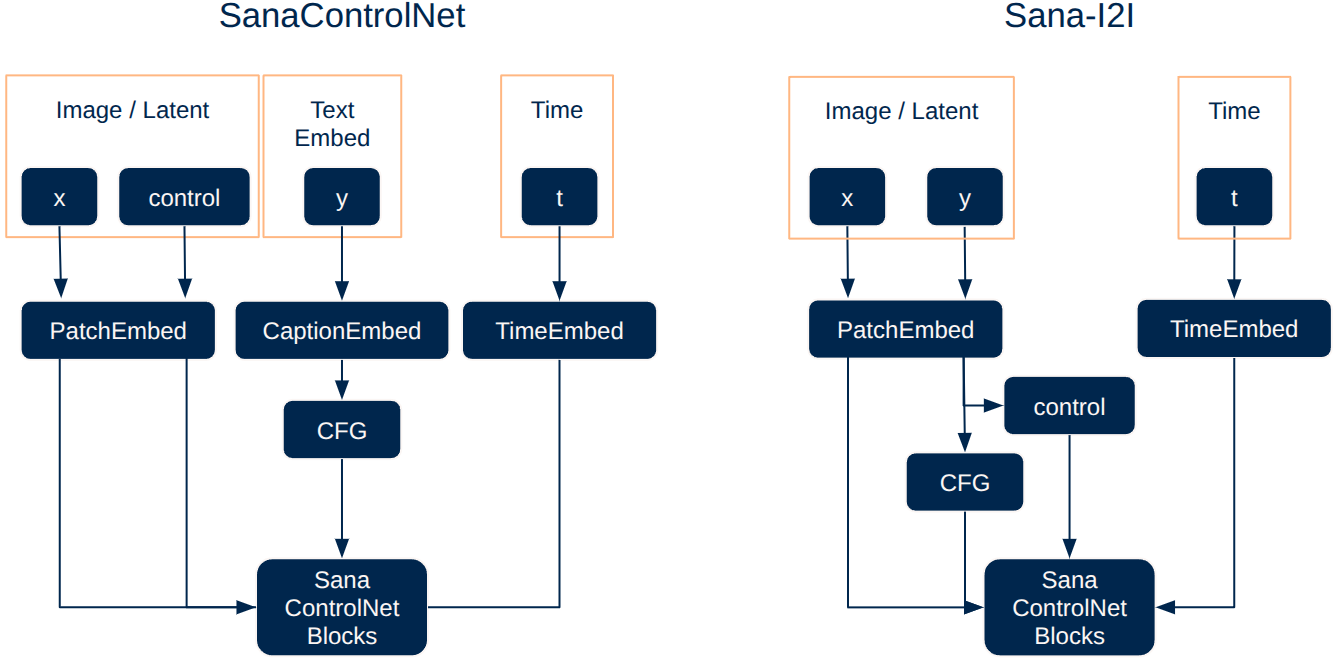}
    \caption{Comparison between the SanaControlNet pipeline and the proposed text-free SANA-I2I architecture.}
    \label{fig:sana-i2i-pipeline}
\end{figure}

A schematic comparison between SanaControlNet and the proposed SANA-I2I architecture is shown in Figure \ref{fig:sana-i2i-pipeline}. In SanaControlNet, classifier-free guidance is enabled by randomly replacing the text embedding $y$ with a null token during training with probability $p_{drop}$. This strategy allows the model to learn both conditional and unconditional branches, enabling guidance control at inference time without retraining. For SANA-I2I, we preserve this CFG mechanism by randomly dropping only the conditioning variable $y$ during training, while always keeping the control features active. Specifically, at each training iteration, we set $y = 0$ with probability $p_{drop}$, while the control pathway remains unchanged. This configuration mirrors the original CFG behavior of SanaControlNet and allows the model to learn conditional and unconditional behaviors in the image-to-image setting.

We additionally explore a variant in which both $y$ and the control signal are simultaneously dropped during training. This enables a fully unconditional generation mode at inference time by setting $y = 0$ and disabling control. However, in the context of fetal artifact removal, this variant yields inferior artifact suppression compared to the primary approach. Consequently, we treat this configuration as exploratory and include its results only for completeness.

\section{Data Description} \label{sec:dados}

The fetal MRI dataset employed in this work is based on the dataset previously introduced in \cite{santos2024fetal}. The original dataset consists of fetal MRI scans from $38$ patients acquired using a $1.5$-T scanner (Magnetom Aera, Siemens, Erlangen, Germany) with a 3D T2-weighted true fast imaging sequence with steady-state precession (trueFISP). As in \cite{santos2024fetal}, we applied a 3D T2-weighted true fast imaging sequence with steady-state precession (truefisp) in sagittal plane (TR/TE = $3.02/1.43$ $ms$); isotropic voxel $1.0 \times 1.0 \times 1.0$ $mm^3$ ); matrix: $256 \times 256$ $mm$, $136$ slices, with a total acquisition time of $26s$. Maternal sedation was not used.

In the previous study, both real images with motion artifacts (domain A) and images acquired under controlled conditions with reduced artifacts (domain B) were used in an unpaired setting. In contrast, the present work requires a \emph{paired} dataset for training. Since real paired fetal MRI images cannot be reliably obtained due to fetal and maternal motion, we exclusively rely on synthetically generated paired data. The paired data are generated by retrospectively simulating motion artifacts on images from domain B using the method proposed by \cite{Duffy2021RetrospectiveMA}, as previously adopted in \cite{santos2024fetal}. This process results in a synthetic corrupted domain (domain C), where each image in domain C is explicitly paired with its corresponding clean image in domain B.

In contrast to our previous work \cite{santos2024fetal}, where extensive stochastic data augmentation was employed to mitigate overfitting, the present study adopts a deterministic preprocessing pipeline during training. This modification was introduced to ensure consistency between synthetic artifact generation and model input characteristics, as well as to reduce variability that could interfere with controlled evaluation. First, images are resized to a predefined spatial resolution using the selected interpolation strategy. A center crop is then applied to ensure spatial consistency across samples. The images are subsequently converted to tensor format and normalized to the range $[-1, 1]$ to stabilize training and improve convergence behavior.

The main difference with respect to \cite{santos2024fetal} lies in the level of artifacts used when generating the synthetic paired data. In the previous work, multiple artifact severities were explored by imposing upper and lower bounds on the Structural Similarity Index Measure (SSIM) \cite{SSIM-Ref2004}, such that a corrupted image $\mathbf{y}$ was accepted only if its similarity with the clean reference $\mathbf{x}$ satisfied
\[
s_{0} < \mathrm{SSIM}(\mathbf{x}, \mathbf{y}) < s_{1}.
\]
These parameters $(s_{0}, s_{1})$ therefore control the \emph{severity range} of the synthetic motion artifacts: lower SSIM values correspond to stronger distortions and more destructive motion degradation, whereas higher SSIM values correspond to milder artifacts.

In the present work, we exclusively adopt the most severe artifact configuration, defined as
\[
(s_{0}, s_{1}) = (0.6,\ 0.9).
\]
This range generates synthetic images exhibiting substantial motion-induced corruption which including large spatial inconsistencies and strong ghosting, thus creating a more challenging reconstruction task. This choice is supported by empirical evidence from prior work, which showed that training with stronger artifacts improves robustness and better equips the model to correct large deviations from the underlying fetal anatomy.

All experiments in this study are therefore conducted using paired synthetic data generated under this SSIM interval. In addition to the synthetic evaluation, we also report results on real motion-degraded fetal MRI, enabling assessment of the method’s performance and generalization to real world acquisition conditions.

\section{Training} \label{sec:training-configs}

The proposed method was trained following the same optimization strategy and hyperparameter configuration adopted in the original SanaControlNet framework \cite{sana2024}. In order to ensure a fair and controlled comparison, all training parameters related to optimization, scheduling, and numerical precision were preserved without modification. The only differences concern the input resolution and the architectural adaptation required for image-only conditioning.

Training was performed for 100 epochs using a batch size of 16 and a single-step gradient accumulation strategy. Mixed-precision training (FP16) was employed to improve computational efficiency, while attention layers were maintained in FP32 for numerical stability. Gradient checkpointing was enabled to reduce memory consumption, and gradient clipping was applied with a maximum norm of 0.1 to stabilize optimization.

We adopted the same optimizer configuration as the original SanaControlNet implementation, using the CAME optimizer \cite{luo2023came} with learning rate $1 \times 10^{-4}$, zero weight decay, and $\beta$ coefficients $(0.9, 0.999, 0.9999)$. The learning rate schedule was constant with a warm-up phase of 30 steps. All experiments were conducted with a fixed random seed of 1 to ensure reproducibility.

The diffusion process followed the linear flow noise schedule with velocity prediction enabled, consistent with the original Sana framework. Logit-normal timestep weighting was used with mean 0.0 and standard deviation 1.0, and inference sampling was performed using the flow DPM-solver strategy with 500 sampling steps during evaluation.

The model was initialized from the pretrained Sana600M checkpoint at 512$\times$512 resolution. Unlike the original SanaControlNet configuration, which operates at 1024$\times$1024 resolution and incorporates explicit control signals, the proposed method operates at 512$\times$512 resolution and employs an image-only control formulation. Although the configuration file retains the \texttt{control\_signal\_type} field for compatibility with the framework, no external control signal was used during training. Additionally, the DC-AE is maintained identical to the original \cite{chen2025dcae}.

By maintaining the original optimization and scheduling parameters, the experimental design ensures that performance differences arise from the proposed methodological modifications rather than changes in training configuration. This controlled setup allows a direct comparison with the baseline SanaControlNet model while adapting the framework to the specific requirements for image-to-image translations tasks.

\section{Results} \label{sec:results}

To assess the effectiveness of the proposed method, we conducted a case study in the context of the motion artifacts reduction of fetal MRI under realistic clinical constraints. The experimental framework was designed steered by the practical limitation that paired corrupted and artifact-free acquisitions are not available.

Three image domains were considered throughout the experiments. Domain A consists of real MR images containing acquisition with artifacts. Domain B comprises real artifact-free MR images acquired under a well-controlled capture protocol. Due to the inherent nature of MR acquisition, it is not possible to obtain paired samples across Domains A and B, as the same anatomical instance cannot be simultaneously acquired with and without artifacts. To enable controlled quantitative assessment, we generated a third dataset, Domain C, by simulating realistic motion artifacts on images from Domain B using the Duffy method \cite{Duffy2021RetrospectiveMA}. This procedure produced paired samples between Domains C (synthetic corrupted) and B (clean ground truth), which were used for model training and synthetic evaluation. 

Accordingly, performance was assessed under two complementary conditions: (i) a synthetic paired evaluation (Domain C $\rightarrow$ Domain B), where full-reference metrics can be computed, and (ii) a real-world unpaired evaluation (Domain A $\rightarrow$ Domain B), where assessment must rely on visual inspection or distribution-based measures due to the absence of pairs. This dual evaluation strategy enables both controlled quantitative validation and realistic performance analysis under clinical conditions. In addition, we also include the results of the Semi-supervised Cycle GAN version proposed in our previous work for comparison purposes \cite{santos2024fetal}.

All experiments performed considered a fixed number of five steps with a guidance scale of 1.0, chosen experimentally to minimize the Fréchet Inception Distance (FID) score \cite{heusel2017fid}. More details are given in the section \ref{sec:ablation-study}, where the effectiveness of the guidance scale and the number of steps on the Sana-I2I are verified.

\subsection{Quantitative Evaluation on Synthetic Data}

In the synthetic setting (Domain C $\rightarrow$ Domain B), reconstruction performance was evaluated using Structural Similarity Index (SSIM) \cite{SSIM-Ref2004} and Mean Absolute Error (MAE). SSIM measures perceptual similarity by assessing structural consistency between reconstructed images and ground truth, while MAE quantifies average pixel-wise reconstruction error. Table~\ref{tab:synthetic_results} summarizes the quantitative results obtained on the synthetic dataset.

\begin{table}[h]
\centering
\caption{Quantitative evaluation on synthetic paired data (Domain C $\rightarrow$ Domain B).}
\label{tab:synthetic_results}
\begin{tabular}{lcc}
\hline
Method & SSIM $\uparrow$ & MAE (normed) $\downarrow$ \\
\hline
Previous Method \cite{santos2024fetal} & \textbf{0.748} & \textbf{0.036} \\
Proposed Method & 0.724 & 0.039 \\
\hline
\end{tabular}
\end{table}

By observing the Table \ref{tab:synthetic_results}, we notice that the SSIM and MAE of the previous method \cite{santos2024fetal} are slight better than the ones achieved for the proposed method. This behavior occurs because when the original image is darker than usual the SANA-I2I tends to improve the luminance while the previous method preserves the original brightness.. Although this increases the pixel-wise error (MAE) and reduces SSIM, it does \emph{not} imply worse reconstruction quality. In fact, as illustrated in Figure \ref{fig:qualitative_results}, the previous method often retains strong residual motion artifacts, whereas the proposed model effectively removes them. Because SSIM and MAE operate on pixel-level comparisons, these metrics penalize luminance correction even when the resulting image is more anatomically meaningful and visually closer to the target distribution.

Thus, the SSIM/MAE values should be interpreted with caution in this context: the proposed method changes global intensity inconsistencies that the previous method leaves untouched, which artificially lowers full-reference metrics while actually producing cleaner and more artifact-free outputs.

\subsection{Evaluation on Real Corrupted Images}

Since real corrupted images from Domain A do not have corresponding clean ground-truth references, full-reference metrics cannot be computed. To evaluate performance under this realistic scenario, we employed distribution-based measures, namely Fréchet Inception Distance (FID) \cite{heusel2017fid} and Kernel Inception Distance (KID) \cite{binkowski2018kid}, to assess how closely the corrected images resemble the distribution of artifact-free images from Domain B.

FID quantifies the distance between feature distributions extracted from two image sets, comparing the mean and covariance of deep feature representations \cite{heusel2017fid}. Kernel Inception Distance (KID), in contrast, computes the squared Maximum Mean Discrepancy (MMD) between embedded distributions without assuming Gaussianity \cite{binkowski2018kid}. Lower values of both metrics indicate better alignment between the restored real images and the clean reference domain. The results are presented in Table~\ref{tab:real_results}.

\begin{table}[h]
\centering
\caption{Distribution-based evaluation on real corrupted images (Domain A).}
\label{tab:real_results}
\begin{tabular}{lcc}
\hline
Method & FID $\downarrow$ & KID $\downarrow$ \\
\hline
Previous Method \cite{santos2024fetal} & 60.14 & 0.031038 $\pm$ 0.000878 \\
Proposed Method & \textbf{51.74} & \textbf{0.023725 $\pm$ 0.000856} \\
\hline
\end{tabular}
\end{table}

Unlike SSIM and MAE, the distribution-based metrics (FID and KID) clearly favor the proposed method. Because these metrics measure similarity at the feature-distribution level rather than through pixel-wise comparisons, they are less affected by global luminance corrections and instead capture improvements in structural realism and artifact suppression. The lower FID and KID scores obtained by the proposed method indicate that its outputs are statistically closer to the distribution of clean fetal MR images.

It is important to note that Domain B, while defined as artifact-free, still contains images with mild residual artifacts, which may influence the absolute values of FID and KID. However, because both methods are compared against the same reference distribution, the improvement observed in FID and KID demonstrates that the proposed technique produces images that more closely match the appearance and statistical characteristics of high-quality fetal MRI.

Overall, while full-reference metrics (SSIM, MAE) may penalize beneficial luminance corrections introduced by the proposed model, the distribution-based evaluation confirms that it generates images that more faithfully align with the clean target domain and better suppress motion artifacts.

\subsection{Qualitative Analysis}

\begin{figure}[t]
    \centering
    
    \includegraphics[width=0.32\textwidth]{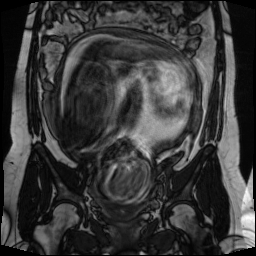}
    \hfill
    \includegraphics[width=0.32\textwidth]{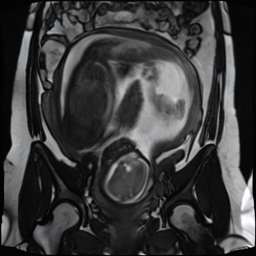}
    \hfill
    \includegraphics[width=0.32\textwidth]{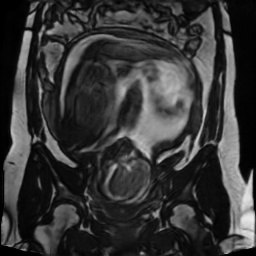}
    
    \vspace{2pt}
    
    \includegraphics[width=0.32\textwidth]{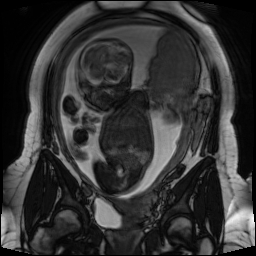}
    \hfill
    \includegraphics[width=0.32\textwidth]{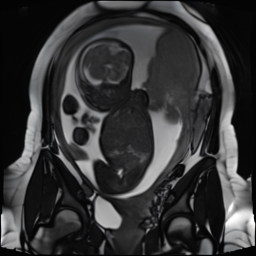}
    \hfill
    \includegraphics[width=0.32\textwidth]{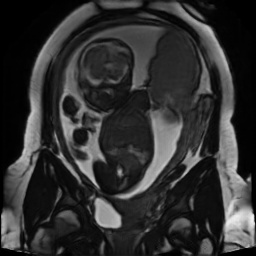}
    
    \vspace{2pt}
    
    \includegraphics[width=0.32\textwidth]{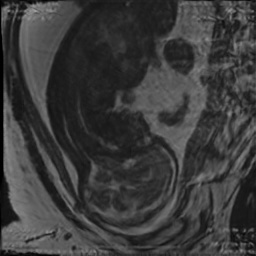}
    \hfill
    \includegraphics[width=0.32\textwidth]{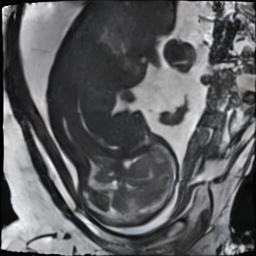}
    \hfill
    \includegraphics[width=0.32\textwidth]{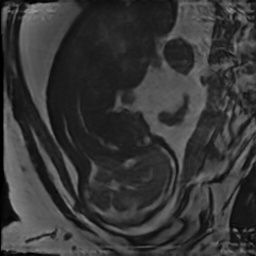}
    
    \vspace{2pt}
    
    \includegraphics[width=0.32\textwidth]{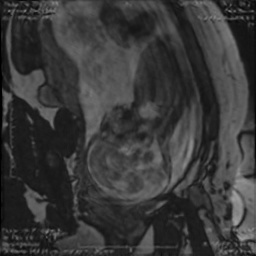}
    \hfill
    \includegraphics[width=0.32\textwidth]{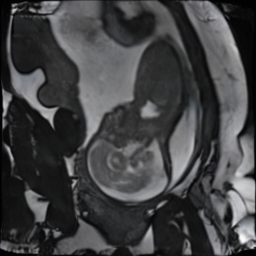}
    \hfill
    \includegraphics[width=0.32\textwidth]{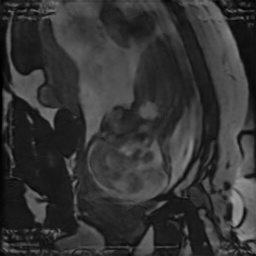}
    
    
    
    
    
    \vspace{2pt}
    
    \begin{tabular}{ccc}
    (a) Original/Input \ \ \ \ \ \ \ \ \ \ \ \ \ \ \ \ \ \ \  \ \ \ \ \ \ \ \ \ &
    (b) Proposed Method \ \ \ \ \ \ \ \ \ \ \ \ \ \ \ \ \ \ \  \ \ \ \ \ \ \ \ \ \ \ \ \ \ &
    (c) CycleGAN V2
    \end{tabular}
    
    \caption{Qualitative comparison between the proposed method and CycleGAN V2. Each row corresponds to a different input image. Columns represent (a) the original input image, (b) the result obtained using the proposed method, and (c) the result produced by CycleGAN V2.}
    \label{fig:qualitative_results}
\end{figure}

To complement the quantitative analysis, we performed a detailed visual inspection of restored real corrupted images from Domain A. Representative examples are shown in Figure \ref{fig:qualitative_results}, where the outputs of the previous method and the proposed technique are compared.
    
Visual assessment reveals that the proposed method achieves more effective artifact suppression while preserving anatomical boundaries and fine structural details. In contrast, the previous approaches tends to leave residual artifact patterns and, in some cases, introduces mild over-smoothing effects. Importantly, the proposed model does not introduce visually implausible structures, suggesting stable behavior when applied to real corrupted data.

\subsection{Ablation Study} \label{sec:ablation-study}

\begin{figure}[t]
    \centering
    
    \begin{subfigure}{0.25\textwidth}
        \centering
        \includegraphics[width=\linewidth]{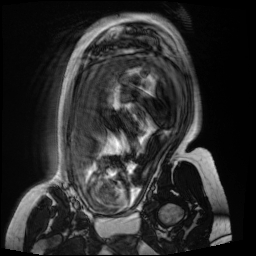}
        \caption{Orginal}
    \end{subfigure}
    \hfill
    \begin{subfigure}{0.25\textwidth}
        \centering
        \includegraphics[width=\linewidth]{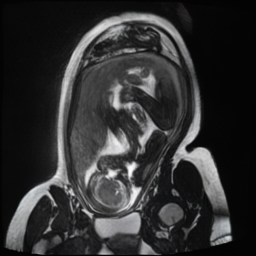}
        \caption{2 steps}
    \end{subfigure}
    \hfill
    \begin{subfigure}{0.25\textwidth}
        \centering
        \includegraphics[width=\linewidth]{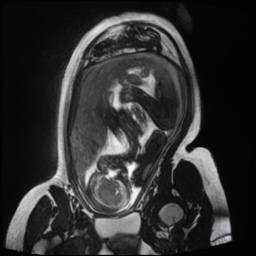}
        \caption{5 steps}
    \end{subfigure}
    \hfill
    \begin{subfigure}{0.25\textwidth}
        \centering
        \includegraphics[width=\linewidth]{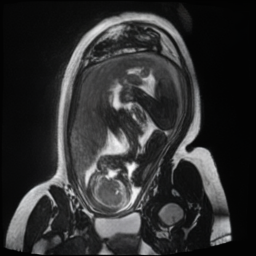}
        \caption{10 steps}
    \end{subfigure}
    \hfill
    \begin{subfigure}{0.25\textwidth}
        \centering
        \includegraphics[width=\linewidth]{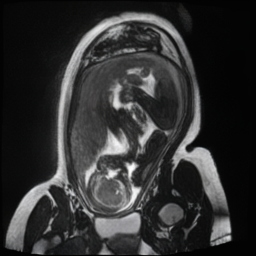}
        \caption{20 steps}
    \end{subfigure}
    \hfill
    \begin{subfigure}{0.25\textwidth}
        \centering
        \includegraphics[width=\linewidth]{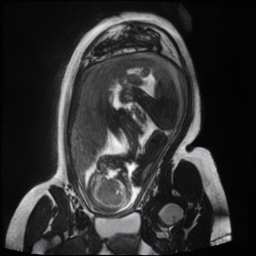}
        \caption{40 steps}
    \end{subfigure}
    
    \caption{Variation of number of steps in $[2, 5, 10, 20, 40]$ with a fixed guidance scale of $1.0$.}
    \label{fig:var-nsteps}

\end{figure}

In this set of experiments, we evaluate the influence of the guidance scale and the number of sampling steps on the resulting images. The first point of interest is determining how many steps are required to produce a satisfactory reconstruction. To assess this in the context of fetal MRI motion artifact reduction, we fixed the guidance scale at $1.0$, a value that consistently produced good-quality outputs, and varied the number of steps in the set $[2, 5, 10, 20, 40]$. One of the corresponding experiments can be visualized in Figure \ref{fig:var-nsteps}. Visual inspection shows that image quality changes very little beyond $5$ steps, and even with only $2$ sampling steps the results were already sufficiently good.

\begin{figure}[H]
    \centering
    
    \begin{subfigure}{0.19\textwidth}
    \centering
    \includegraphics[width=\linewidth]{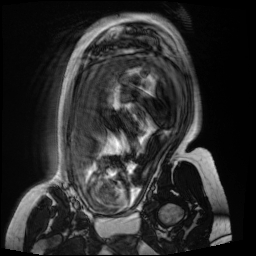}
    \caption{Original}
    \end{subfigure}
    
    \vspace{6pt}
    
    \begin{subfigure}{0.19\textwidth}
    \centering
    \includegraphics[width=\linewidth]{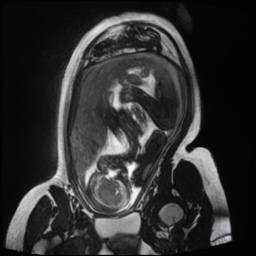}
    \caption{Guidance scale 1.0}
    \end{subfigure}
    \hfill
    \begin{subfigure}{0.19\textwidth}
    \centering
    \includegraphics[width=\linewidth]{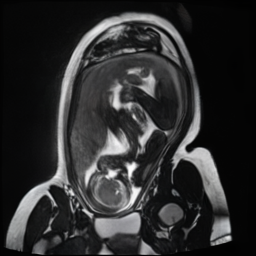}
    \caption{Guidance scale 1.1}
    \end{subfigure}
    \hfill
    \begin{subfigure}{0.19\textwidth}
    \centering
    \includegraphics[width=\linewidth]{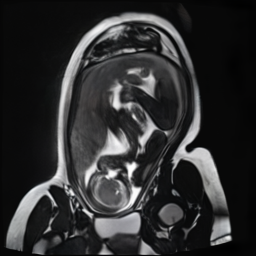}
    \caption{Guidance scale 1.2}
    \end{subfigure}
    \hfill
    \begin{subfigure}{0.19\textwidth}
    \centering
    \includegraphics[width=\linewidth]{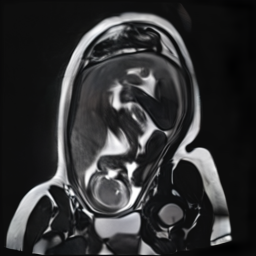}
    \caption{Guidance scale 1.3}
    \end{subfigure}
    \hfill
    \begin{subfigure}{0.19\textwidth}
    \centering
    \includegraphics[width=\linewidth]{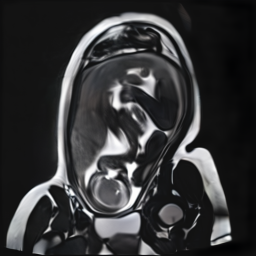}
    \caption{Guidance scale 1.4}
    \end{subfigure}
    
    \vspace{6pt}
    
    \begin{subfigure}{0.19\textwidth}
    \centering
    \includegraphics[width=\linewidth]{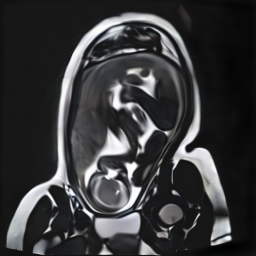}
    \caption{Guidance scale 1.5}
    \end{subfigure}
    \hfill
    \begin{subfigure}{0.19\textwidth}
    \centering
    \includegraphics[width=\linewidth]{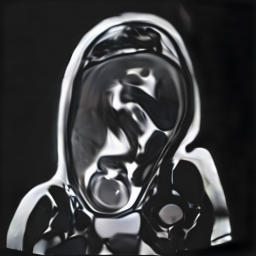}
    \caption{Guidance scale 1.6}
    \end{subfigure}
    \hfill
    \begin{subfigure}{0.19\textwidth}
    \centering
    \includegraphics[width=\linewidth]{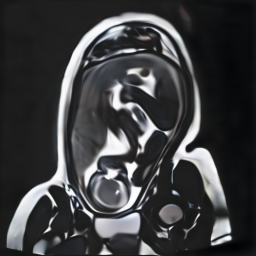}
    \caption{Guidance scale 1.7}
    \end{subfigure}
    \hfill
    \begin{subfigure}{0.19\textwidth}
    \centering
    \includegraphics[width=\linewidth]{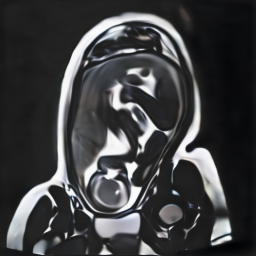}
    \caption{Guidance scale 1.8}
    \end{subfigure}
    \hfill
    \begin{subfigure}{0.19\textwidth}
    \centering
    \includegraphics[width=\linewidth]{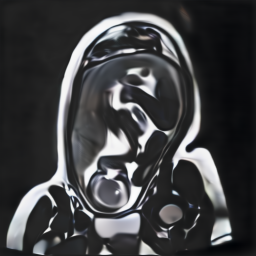}
    \caption{Guidance scale 1.9}
    \end{subfigure}
    
    \caption{Variation of guidance scale in the interval $(1.0, 1.9)$ with a fixed number of steps.}
    \label{fig:var-guidance}

\end{figure}

The second point concerns the influence of the guidance scale on the generated images. For this experiment, we fixed the number of sampling steps to $5$ (as supported by the previous analysis) and varied the guidance scale in the interval $(1.0, 1.9)$ using a set of reference images. One of these experiments is shown in Figure \ref{fig:var-guidance}.

Visual inspection of the results indicates that increasing the guidance scale leads to more aggressive artifact suppression. However, this also results in the loss of important anatomical details relevant for diagnosis. In this regard, the most suitable values for the guidance scale were between $1.0$ and $1.1$. An evaluation based on the FID score confirmed that a guidance scale of $1.0$ achieved the best FID values when comparing generated images from domain A (real artifact images) with domain B (artifact-free images).

\begin{figure}[H]
    \centering
    
    \begin{subfigure}{0.19\textwidth}
        \centering
        \includegraphics[width=\linewidth]{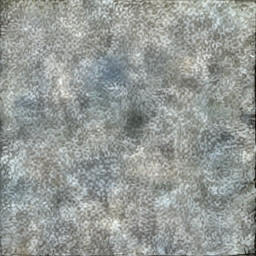}
        \caption{2 steps}
        \end{subfigure}
        \hfill
        \begin{subfigure}{0.19\textwidth}
        \centering
        \includegraphics[width=\linewidth]{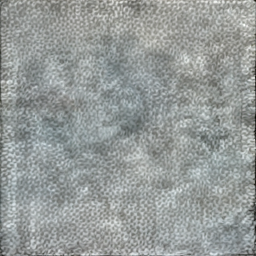}
        \caption{5 steps}
        \end{subfigure}
        \hfill
        \begin{subfigure}{0.19\textwidth}
        \centering
        \includegraphics[width=\linewidth]{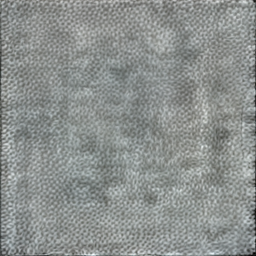}
        \caption{10 steps}
        \end{subfigure}
        \hfill
        \begin{subfigure}{0.19\textwidth}
        \centering
        \includegraphics[width=\linewidth]{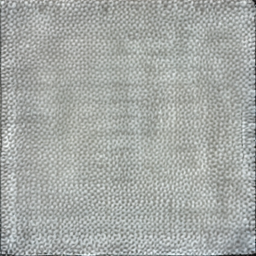}
        \caption{20 steps}
        \end{subfigure}
        \hfill
        \begin{subfigure}{0.19\textwidth}
        \centering
        \includegraphics[width=\linewidth]{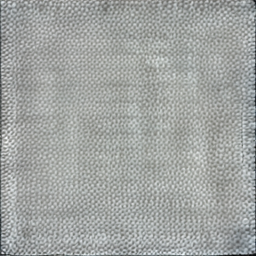}
        \caption{40 steps}
    \end{subfigure}
    
    \caption*{(a)}
    
    \vspace{6pt}
    
    \begin{subfigure}{0.19\textwidth}
        \centering
        \includegraphics[width=\linewidth]{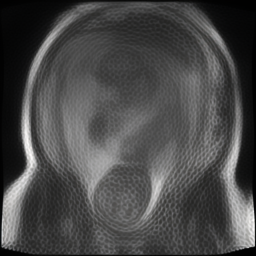}
        \caption{2 steps}
        \end{subfigure}
        \hfill
        \begin{subfigure}{0.19\textwidth}
        \centering
        \includegraphics[width=\linewidth]{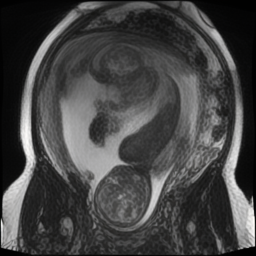}
        \caption{5 steps}
        \end{subfigure}
        \hfill
        \begin{subfigure}{0.19\textwidth}
        \centering
        \includegraphics[width=\linewidth]{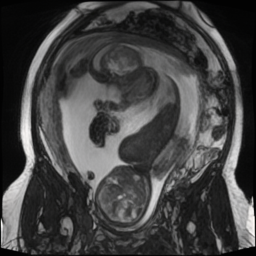}
        \caption{10 steps}
        \end{subfigure}
        \hfill
        \begin{subfigure}{0.19\textwidth}
        \centering
        \includegraphics[width=\linewidth]{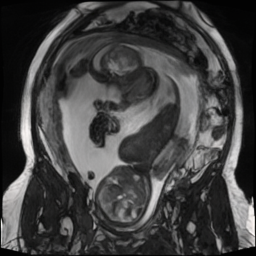}
        \caption{20 steps}
        \end{subfigure}
        \hfill
        \begin{subfigure}{0.19\textwidth}
        \centering
        \includegraphics[width=\linewidth]{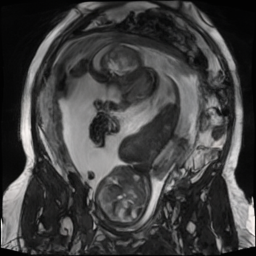}
        \caption{40 steps}
    \end{subfigure}
    
    \caption*{(b)}
    
    \caption{(a) Image generation using guidance scale $0$ with the proposed method. (b) Image generation using guidance scale $0$ with the BIS method.}
    \label{fig:gen-images-g0}

\end{figure}

In Section \ref{sec:methodology}, we presented an alternative method (BIS) that supports both image correction with guidance scales greater than zero and unconditional image generation when the guidance scale is set to zero. However, for fetal image restoration, the BIS method produced inferior results when compared to the primary approach. Nevertheless, for completeness, we report an additional experiment in which we generated images using a guidance scale of $0$ and varied the number of sampling steps in $[2, 5, 10, 20, 40]$ for both methods, that is, Primary and BIS approaches. As illustrated in Figure \ref{fig:gen-images-g0}, the primary method, which yields the best results for fetal image correction, is unable to generate images when the guidance scale is zero, whereas the BIS method is capable of producing reasonable images when using at least 10-20 steps.

\section{Final Remarks} \label{sec:final-remarks}

The results presented in this work demonstrate that the proposed text-free, flow‑matching image-to-image framework is effective for paired reconstruction tasks. Although fetal MRI was used as the initial case study, the methodology itself is general and can be applied to any paired domain. In the synthetic paired setting, the proposed method achieves a level of artifact suppression that is visually superior to the previous approach, even though SSIM and MAE do not always reflect this improvement. As discussed earlier, these full-reference metrics are sensitive to global luminance adjustments, which the proposed model actively corrects. Consequently, SSIM and MAE may penalize beneficial intensity normalization, despite the resulting images exhibiting fewer motion artifacts and clearer anatomical structures.

In contrast, the distribution-based metrics computed on real corrupted data, FID and KID provide a more reliable indication of perceptual and statistical fidelity. The consistent reduction in both metrics confirms that the proposed method produces outputs that lie closer to the distribution of high-quality fetal MR images. This improvement is particularly meaningful because Domain B, although considered artifact-free, still contains mild residual artifacts. Nevertheless, the proposed model achieves better alignment with this reference distribution than the previous method.

The qualitative evaluation reinforces these findings: the proposed method suppresses motion artifacts more effectively and preserves structural boundaries without introducing implausible anatomical features. Additionally, the ablation study provides important insights into the model’s behavior. In particular, the experiments show that a small number of sampling steps is sufficient to achieve high-quality reconstructions, and that moderate guidance scales (between 1.0 and 1.1) offer the best trade-off between artifact removal and anatomical detail preservation. While the BIS variant is able to generate unconditional samples at zero guidance scale, it remains less effective for artifact correction in fetal MRI when compared to the Primary approach SANA-I2I.

Overall, the combination of quantitative results, qualitative inspection, and ablation studies highlights the robustness, efficiency, and broader applicability of the proposed approach beyond the specific fetal MRI setting explored in this study.

\bibliographystyle{plain}
\bibliography{references}  

@article{song2020score,
  title={Score-Based Generative Modeling through Stochastic Differential Equations},
  author={Song, Yang and sohl-Dickstein, Sohl and Kingma Diederik and Kumar, Abhishek and Poole, Ben and Ermon, Stefano},
  journal={ICLR},
  year={2021}
}

@inproceedings{dhariwal2021diffusion,
  title={Diffusion Models Beat GANs on Image Synthesis},
  author={Dhariwal, Prafulla and Nichol, Alex},
  booktitle={NeurIPS},
  year={2021}
}

@article{lipman2023flow,
  title={Flow Matching for Generative Modeling},
  author={Lipman, Yaron and Chen, Ricky and Ben-Hamu, Heli and Nickel, Maximilian and Le, Matt},
  journal={ICLR},
  year={2023}
}

@article{sana2024,
  title={{SANA}: Efficient High-Resolution Flow Matching for Scalable Image Synthesis},
  author={Xie, Enze and Chen, Junsong and Chen, Junyu and Cai, Han and Tang, Haotian and Lin, Yujun and Zhang, Zhekai and Li, Muyang and Zhu, Ligeng and Lu, Yao and Han, Song},
  journal={arXiv},
  year={2024}
}

@inproceedings{zhang2023controlnet,
  title={Adding Conditional Control to Text-to-Image Diffusion Models},
  author={Zhang, Lvmin and Rao, Anyi and Agrawala, Maneesh},
  booktitle={ICCV},
  year={2023}
}

@inproceedings{isola2017pix2pix,
  title={Image-to-Image Translation with Conditional Adversarial Networks},
  author={Isola, Phillip and Zhu, Jun-Yan and Zhou, Tinghui and Efros, Alexei},
  booktitle={CVPR},
  year={2017}
}

@article{zhang2017dncnn,
  title={Beyond a Gaussian Denoiser: Residual Learning of Deep CNN for Image Denoising},
  author={Zhang, Kai and Zuo, Wangmeng and Chen, Yunjin and Meng, Deyu and Zhang, Lei},
  journal={IEEE},
  year={2017}
}

@article{chartsias2017multimodal,
  title={Multimodal MRI Synthesis with Generative Adversarial Networks},
  author={Chartsias, Agisilaos and Joyce, Thomas},
  journal={ISBI},
  year={2017}
}

@article{xia2023diffi2i,
  title={DiffI2I: Efficient Diffusion Model for Image-to-Image Translation},
  author={Xia, Bin and Zhang, Yulun and Wang, Shiyin and Wang, Yitong and Wu, Xinglong and Tian, Yapeng and Yang, Wenming and Timofte, Radu and Van Gool, Luc},
  journal={arXiv},
  year={2023},
}

@inproceedings{li2023bbdm,
  title={BBDM: Image-to-Image Translation with Brownian Bridge Diffusion Models},
  author={Li, Bo and Xue, Kaitao and Liu, Bin and Lai, Yu-Kun},
  booktitle={CVPR},
  year={2023},
}

@article{xia2024dmt,
  title={A Diffusion Model Translator for Efficient Image-to-Image Translation},
  author={Xia, Mengfei and Zhou, Yu and Yi, Ran and Liu, Yong-Jin and Wang, Wenping},
  journal={IEEE Transactions on Pattern Analysis and Machine Intelligence},
  volume={46},
  number={12},
  pages={10272-10283},
  year={2024},
  doi={10.1109/TPAMI.2024.3435448},
}

@article{saharia2022palette,
  title={Palette: Image-to-Image Diffusion Models},
  author={Saharia, Chitwan and Chan, William and Chang, Huiwen and Lee, Chris and Ho, Jonathan and Salimans, Tim and Fleet, David and Norouzi, Mohammad},
  journal={ACM SIGGRAPH 2022 Conference Proceedings},
  year={2022}
}

@article{chadebec2025lbm,
  title={LBM: Latent Bridge Matching for Fast Image-to-Image Translation},
  author={Chadebec, Cl{\'e}ment and Ta{\c{s}}ar, Onur and Sreetharan, Sanjeev and Aubin, Benjamin},
  journal={arXiv},
  year={2025}
}

@article{ho2022classifierfree,
  title={Classifier-Free Diffusion Guidance},
  author={Ho, Jonathan and Salimans, Tim},
  journal={NeurIPS Workshop},
  year={2021}
}

@inproceedings{luo2023came,
  title={CAME: Confidence-guided Adaptive Memory Efficient Optimization},
  author={Luo, Yang and Ren, Xiaozhe and Zheng, Zangwei and Jiang, Zhuo and Jiang, Xin and You, Yang},
  booktitle={Proceedings of the 61st Annual Meeting of the Association for Computational Linguistics (ACL)},
  pages={4442--4453},
  year={2023},
  doi={10.18653/v1/2023.acl-long.243},
}

@inproceedings{heusel2017fid,
  title={GANs Trained by a Two Time-Scale Update Rule Converge to a Local Nash Equilibrium},
  author={Heusel, Martin and Ramsauer, Hubert and Unterthiner, Thomas and Nessler, Bernhard and Hochreiter, Sepp},
  booktitle={Advances in Neural Information Processing Systems (NeurIPS)},
  year={2017},
}

@inproceedings{binkowski2018kid,
  title={Demystifying MMD GANs},
  author={Bińkowski, Mikołaj and Sutherland, Danica J. and Arbel, Michael and Gretton, Arthur},
  booktitle={International Conference on Learning Representations (ICLR)},
  year={2018},
}

@article{ho2020ddpm,
  title={Denoising Diffusion Probabilistic Models},
  author={Ho, Jonathan and Jain, Ajay and Abbeel, Pieter},
  journal={NeurIPS},
  year={2020}
}

@inproceedings{chen2025dcae,
  title={Deep Compression Autoencoder for Efficient High-Resolution Diffusion Models},
  author={Chen, Junyu and Cai, Han and Chen, Junsong and Xie, Enze and Yang, Shang and Tang, Haotian and Li, Muyang and Lu, Yao and Han, Song},
  booktitle={International Conference on Learning Representations (ICLR)},
  year={2025},
  url={https://arxiv.org/abs/2410.10733}
}

@article{Duffy2021RetrospectiveMA,
    title={Retrospective motion artifact correction of structural MRI images using deep learning improves the quality of cortical surface reconstructions}, 
    author={Duffy, Ben A and Zhao, Lu and Sepehrband, Farshid and Min, Joyce and Wang, Danny JJ and Shi, Yonggang and Toga, Arthur W and Kim, Hosung and Alzheimer's Disease Neuroimaging Initiative and others}, 
    journal={Neuroimage}, 
    volume={230},
    year={2021}, 
    publisher={Elsevier} 
}

@article{Lim2023MotionAC,
title = {Motion artifact correction in fetal MRI based on a Generative Adversarial network method},
journal = {Biomedical Signal Processing and Control},
volume = {81},
pages = {104484},
year = {2023},
issn = {1746-8094},
doi = {https://doi.org/10.1016/j.bspc.2022.104484},
author = {Adam Lim and Justin Lo and Matthias W. Wagner and Birgit Ertl-Wagner and Dafna Sussman}
}

@ARTICLE{SSIM-Ref2004,
  author={Zhou Wang and Bovik, A.C. and Sheikh, H.R. and Simoncelli, E.P.},
  journal={IEEE Transactions on Image Processing}, 
  title={Image quality assessment: from error visibility to structural similarity}, 
  year={2004},
  volume={13},
  number={4},
  pages={600-612},
  doi={10.1109/TIP.2003.819861}
}

@article{santos2024fetal,
  title={Fetal MRI Artifacts: Semi-Supervised Generative Adversarial Neural Network for Motion Artifacts Reducing in Fetal Magnetic Resonance Images},
  author={Santos, Ítalo and Giraldi, Gilson and Werner, Heron and Schulze, Bruno},
  journal={Journal of Computer and Communications},
  volume={12},
  number={6},
  year={2024},
  doi={10.4236/jcc.2024.126013}
}

@article{RetroMRMotionCorrection,
    author = {Spieker, Veronika and Eichhorn, Hannah and Hammernik, Kerstin and Rueckert, Daniel and Preibisch, Christine and Karampinos, Dimitrios and Schnabel, Julia},
    year = {2023},
    month = {10},
    pages = {1-1},
    title = {Deep Learning for Retrospective Motion Correction in MRI: A Comprehensive Review},
    volume = {PP},
    journal = {IEEE Transactions on Medical Imaging},
    doi = {10.1109/TMI.2023.3323215}
}






\end{document}